\documentclass[11pt,a4paper]{article}
\usepackage{authblk}
\usepackage[hyperref]{emnlp2019}
\usepackage{times}
\usepackage[english]{babel}
\usepackage{latexsym}
\usepackage{easy-todo}
\usepackage{url}

\usepackage{tikz}
\usetikzlibrary{positioning, decorations.pathreplacing, matrix}

\aclfinalcopy 

\title{Exploiting Language Instructions for Interpretable and Compositional Reinforcement Learning}
\makeatletter
\newcommand\email[2][]%
   {\newaffiltrue\let\AB@blk@and\AB@pand
      \if\relax#1\relax\def\AB@note{\AB@thenote}\else\def\AB@note{\relax}%
        \setcounter{Maxaffil}{0}\fi
      \begingroup
        \let\protect\@unexpandable@protect
        \def\thanks{\protect\thanks}\def\footnote{\protect\footnote}%
        \@temptokena=\expandafter{\AB@authors}%
        {\def\\{\protect\\\protect\Affilfont}\xdef\AB@temp{#2}}%
         \xdef\AB@authors{\the\@temptokena\AB@las\AB@au@str
         \protect\\[\affilsep]\protect\Affilfont\AB@temp}%
         \gdef\AB@las{}\gdef\AB@au@str{}%
        {\def\\{, \ignorespaces}\xdef\AB@temp{#2}}%
        \@temptokena=\expandafter{\AB@affillist}%
        \xdef\AB@affillist{\the\@temptokena \AB@affilsep
          \AB@affilnote{}\protect\Affilfont\AB@temp}%
      \endgroup
       \let\AB@affilsep\AB@affilsepx
}
\makeatother

\author[*]{Michiel van der Meer}
\author[**]{Matteo Pirotta}
\author[***]{Elia Bruni}
\affil[*]{University of Amsterdam}
\email{\url{{michiel.vandermeer@student.uva.nl}}}
\affil[**]{Facebook AI Research}
\email{\url{{matteo.pirotta@gmail.com}}}
\affil[***]{Universitat Pompeu Fabra}
\email{\url{{elia.bruni@upf.edu}}}

\date{}

\begin{document}
\maketitle

\begin{abstract}
In this work, we present an alternative approach to making an agent compositional through the use of a diagnostic classifier. Because of the need for explainable agents in automated decision processes, we attempt to interpret the latent space from an RL agent to identify its current objective in a complex language instruction. Results show that the classification process causes changes in the hidden states which makes them more easily interpretable, but also causes a shift in zero-shot performance to novel instructions. Lastly, we limit the supervisory signal on the classification, and observe a similar but less notable effect.	
\end{abstract}

\section{Introduction}
As AI becomes more widespread in the real world, and the strive towards universal AI gains more traction, the need for interpretable and general agents increases. Since more systems are performing automated decisions, humans require those systems to explain their behavior, and expect them to work in unknown scenarios.


In this paper, we investigate whether it is possible to train a Reinforcement Learning (RL) agent to operate in a virtual environment while being interpretable in its `intentions', and how its interpretability helps in finding more compositional solutions. More specifically, while training a neural agent to follow some navigation instructions, we require it to spell out what is, at each time-step, its current objective.  To accomplish that, we use the recently introduced \textit{diagnostic classifier}~\citep{hupkes2018visualisation}, a linear classifier which assesses the presence of some specific information in a neural network by trying to predict it from its hidden states. In our case, we use it at training time to predict the current objective of the RL agent.

Our approach is inspired by how humans learn. While
in traditional RL, the objective is defined in terms of a single goal, expressed through some reward function~\citep{sutton2018reinforcement}, when we teach humans to follow instructions, not only do we check for accurate execution, but we also make sure that the instruction, usually expressed in natural language, is correctly understood. Are all word meanings in the instruction known? Is it clear how to segment the instruction such that it can be decomposed in sub-tasks, encouraging efficient sub-task separation~\citep{gopalan2017planning}? In this paper, we account for some of this extra supervision and measure its impact on learning efficiency.
%
%

\section{Related Work}
\label{sec:related_work}

\subsection{Following language instructions}
One of the first attempts at following language instructions is SHRDLU~\citep{winograd1971procedures}. It was designed to understand natural language by relating to a physical world. However, its apparent success stemmed from handwritten rules in a finite grammar, which is unsustainable in natural language. In attempts to deal with incomplete information, probabilistic methods extract cues from the instruction to improve the agent's learning capabilities~\citep{kollar2010toward, vogel2010learning, tellex2011understanding, dzifcak2009and}.

Recently, following language instructions has been actively researched, with the introduction of artificial environments~\citep{bisk2016natural, hermann2017grounded, wu2018building}. Now, the trend has been leaning towards deep reinforcement learning agents, hoping to fullfill the promise of generalizable agents that exploit the instruction~\citep{misra2017mapping, bahdanau2018learning, yu2018interactive}.

We aim to recover a more-human like learning environment; as humans provide linguistic and non-linguistic cues about how to segment instructions (e.g., during execution, by asking the learner to explain some of her actions), we probe the artificial learner for its focus using information from the language instruction.

\subsection{Compositionality}
In the abstract information that an instruction in natural language provides for humans~\citep{werning2012oxford}, artificial agents intuitively should also be able to benefit from the compositionality of language. If the agent is instructed to perform an action on an object that it has never seen before, but does know how to execute the action, it could reuse its knowledge and require less training before it can successfully complete the instruction.

In the context of following navigation commands, \citet{DBLP:conf/icml/LakeB18} introduced the SCAN task, which is designed to test for compositional abilities in neural networks. The authors show how sequence-to-sequence models are generally able to learn navigation commands, but, as soon as they are tested on instructions which require compositional generalization, they fail miserably.


Additionally, intrinsic motivation aids in scaling RL agents through the use of an internal supervisory signal, representing a reward from performing ``interesting'' actions~\citep[e.g.,][]{chentanez2005intrinsically,csimcsek2006intrinsic}. These signals, obtained during unsupervised traversal of the environment, are used to help the agent form a set of skills by exploration.  Later, they can be reused and employed when optimizing for a task. It has been successfully applied in cases such as efficient learning with sparse rewards~\citep{pathak2017curiosity}, or in the development of an embodied robotics actor~\citep{frank2014curiosity}.

Finally, compositionality is also at the core of curriculum learning for RL~\citep[e.g.,][]{NarvekarSS17,FlorensaHWZA17}. The idea is to design and solve a sequence of tasks with increasing complexity and reuse the skills acquired in these task to solve the target task. However, designing the curricula may be as hard as (or even more complex) than directly solving the target tasks (when prior knowledge is unavailable). It is crucial that the appriopriate design is chosen, but experiments show that curriculum learning can be beneficial in scaling the training of RL agents~\citep{wu2016training, gupta2017cooperative}.

Our approach can be seen as an instance of curriculum learning where the prior knowledge is the task instruction and the curriculum leverages the sequential structure of the task.

\subsection{Understanding black box models}
Presently, deep neural networks are mostly black boxes, and creating an understanding of their internal mechanisms remains a shot in the dark. Fortunately, recent work in explainable AI (XAI) attempts to increase the transparency of these models. An overview by \citet{biran2017explanation} distinguishes between two notions of explainability: \emph{justification} and \emph{interpretability}.
Justifications are reasons for decisions an agent might make, but are not necessarily connected to the workings of the agent itself. This means they can be generated for non-interpretable systems, and require no retraining of the original model. Interpretations on the other hand, are inherent to the agent, and should reflect how the agent arrived at its decision through its interal workings.

Recent developments for generating interpretations include using t-SNE plots to visualise the latent space of agents~\citep{zahavy2016graying, jaderberg2018human}, examining the attention patterns when agents make decisions~\citep{greydanus2017visualizing}, including a human in the loop to help a model's interpretability~\citep{lage2018human} and using 'diagnostic classifiers' to decode which specific information is encoded in the network~\citep{hupkes2018visualisation}.

In this work, we encourage the agent to develop a more interpretable policy, which, at any time step, is able to report its current objective. Additionally, we investigate the compositionality after training the interpretable policy.

\section{Approach}
\label{sec:approach}
This section describes the environment, model and setup used in the experiments.

\subsection{BabyAI game}
As a testbed for the learning process we make use of the BabyAI platform~\citep{chevalier2018babyai}, which consists of a grid world environment in which the agent is presented with a structured language instruction. The platform contains different levels, which increase in complexity through a combination of distractors, composite instructions, and sparse rewards. The observation presented to the agent is a 7x7 grid, a 2D representation of the agent's surroundings. This ego-centric view contains a symbolic representation of objects, walls, doors and their colors. The agent has access to actions such as picking up objects and walking around. The compact representation of the grid world allows for fast processing of the observations.

The instruction is given in the Baby Language, a well-defined subset of the English language, which is simple yet diverse.
For all our experiments, we develop customized levels which spin off from the original \textit{GoTo} level. We choose \textit{GoTo} because is the least complex instruction and therefore easiest to learn. The atomic instruction is formed by selecting a color and object type at random, specifying a target for the agent. Optionally, the modifier \emph{twice} or \emph{thrice} can be added to an atomic instuction, much like the SCAN dataset~\citep{lake2017generalization}. Example atomic instruction include \emph{go to the red ball}, \emph{go to a blue box twice} and \emph{go to the yellow key thrice}.

In the case an agent is instructed to visit an object multiple times, upon arriving at a target object the objects are shuffled around the environment. The agent has to visit the same object respectively one or two more times in order to complete the instruction correctly. To prevent infinite length episodes, every instruction has an associated maximum number of steps, corresponding to the complexity of the instruction.

Atomic instructions are subsequently combined through the use of various \emph{task connectors}. By means of these operators a compound instruction $c_{compound}$ can be made consisting of atomic instructions $c_A$ and $c_B$. We consider the following task connectors:
\begin{itemize}
    \item \textbf{Before:} Complete $c_A$ before completing $c_B$. If the agent completes instruction $c_B$ first, the compound instruction fails, and no reward is given.
    \item \textbf{After:} Complete $c_A$ after completing $c_B$. If the agent completes instruction $c_A$ first, the compound instruction fails and no reward is given.
\end{itemize}
Besides combining atomic instructions, the connectors apply to complex instructions as well. In this case, the connectors are left-associative.

\begin{figure}[h]
    \centering
    \begin{tikzpicture}[
    squarednode/.style={rectangle},
    minnode/.style={rectangle, draw=red!0, fill=black!5, thick, minimum size=12mm},
    inputnode/.style={rectangle, draw=blue!60, fill=blue!5, thick, minimum size=12mm},
    ]

    \node[squarednode]      (overview)                          {\includegraphics[width=0.4\textwidth]{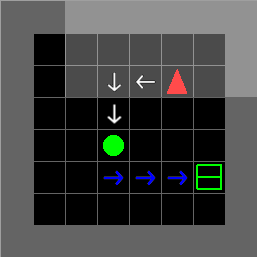}};
    \node[minnode]          (instr)[below=0.1cm of overview]        {\emph{``go to the green ball before go to the green box''}};


    \node (brAstart) [below right=0.1cm and 0.1cm of instr.west]{};
    \node (brAend) [below right=0.1cm and 3.3cm of instr.west]{};
    \draw [
        thick,
        decoration={
            brace,
            mirror,
            raise=0.5cm
        },
        decorate
    ] (brAstart) -- (brAend)
    node [pos=0.5,anchor=north,yshift=-0.55cm,align=center] {$c_A$};

    \node (brBstart) [below left=0.1cm and 0.1cm of instr.east]{};
    \node (brBend) [below left=0.1cm and 3.3cm of instr.east]{};
    \draw [
        thick,
        decoration={
            brace,
            raise=0.5cm
        },
        decorate
    ] (brBstart) -- (brBend)
    node [pos=0.5,anchor=north,yshift=-0.55cm] {$c_B$};
    \end{tikzpicture}

    \caption{Visual overview of the environment. The light gray area is currently in view for the agent, represented by the red triangle. The green ball and green square are the two objectives. Every small arrow is a future action taken by the agent, while simulataneously providing an object classification. For the white arrows, the correct label is ``green ball.'' For the blue arrows, the correct label is ``green box.''}
    \label{fig:setup}
\end{figure}

For example, a compound instruction is \emph{go to the blue box twice before go to the yellow key}. An overview of all levels considered in this work is given in Table~\ref{tab:levels}, and a visual example of the setup is given in Figure~\ref{fig:setup}.

\begin{table*}[h]
    \caption{Overview of all levels. The last column denotes any special feature in each level.}
    \label{tab:levels}
    \centering
    \begin{tabular}{llcc}
        \hline
        \textbf{Level name} & \textbf{Connectors} & \textbf{Num. targets} & \textbf{Other} \\
        \hline
        \textsc{Before} & Before & 2 & None \\
        \textsc{Mixed-2} & Before,After & 2 & None \\
        \textsc{Before (repeat)}  & Before & 2 & Twice/Thrice modifier \\
        \textsc{Mixed-3} & Before,After & 3 & None \\
        \hline
    \end{tabular}
\end{table*}

\subsection{Model}
For our base agent, we select the Small BabyAI model, originally introduced by~\citet{chevalier2018babyai}. This model combines the language instruction and world representation in an Action-Critic architecture~\citep{szepesvari2010algorithms}. The instruction is parsed using a GRU using a fixed vocabulary, after which it is combined with the observation through two FiLM~\citep{perez2018film} layers. The output generated by these layers is passed into an LSTM to allow for temporal feedback connections. Ultimately, the LSTM's output is used in an actor network to generate actions and a critic network to generate state values. The agent is optimized using Proximal Policy Optimization~\citep[PPO, ][]{schulman2017proximal}, a sample efficient actor-critic approach.

\subsection{Diagnostic classification}
As an extension to the base model, the model is made \emph{interpretable} through the addition of a diagnostic classifier. This classifier is tasked with providing an intuitive explanation of the agent's behavior when asked, making it more interpretable for humans. It does so by generating, at every time step, the current target for the agent. While this does not directly give a justification for individual moves, it does give an idea of the current \emph{focus} of the agent. Since we consider complex instructions, there are at least two subtasks to be completed, and through the classification the agent signifies its current objective (e.g. I'm trying to complete $c_A$). By means of this extra task, we aim to make the agent \emph{aware} of the compositional nature of the instruction. The agent now has access to a signal that indicates the separation between two objects in its environment, and it is up to the agent to learn to compose previously learned behavior, and become more efficient.

To create the labels for the diagnostic classifier, we exploit the temporal relation between the subtasks. This way, the agent is trained to visit the objectives in order, and the focus of the agent should follow this same order. In the levels, there are $N$ unique object type/color combinations, as can be generated by the Baby Language. By enumerating all $N$ combinations, a mapping can be created. Subsequently, the labeler takes the language instruction $c_{compound}$ and the current status of visits (e.g. whether the agent has visited $c_A$), and uses this mapping to generate a label. Since only the final label, and not the grammar or task status is exposed to the agent, we avoid providing further external information.

Finally, the labels are used to train the diagnostic classifier, which is a linear mapping from the LSTM's hidden state to the $N$ unique object/color combinations. Because of the classification task, a cross entropy term is added to the PPO reward function $\mathcal{L}^{PPO}$ with a coefficient $\beta$. This results in Equation~\ref{eq:reward}, which takes class labels $y_c$ and output probabilities $p_c$.

\begin{equation}
    \label{eq:reward}
    \mathcal{L} = \mathcal{L}^{PPO} - \beta \sum^{C}_{c=1}y_c \log(p_c)
\end{equation}
Note that this differs from regularized approaches for RL where the regularization term is computed w.r.t.\ the current policy estimate~\citep[e.g.,][]{neu2017unified}. This regularization term can be interpreted as a form of reward shaping~\citep{ng1999policy}.

\section{Experiments}
\label{sec:experiments}

Below, four experiments are outlined. The experiments are designed to quantify how the additional classifier in the agent is affecting its interpretability, and to check whether it has impacted the agent's compositionality.

As a measure of the agent's performance over time, different metrics are used. These metrics show how proficient an agent is in completing the overall instruction, or how consistently it can complete levels. The following are used:

\begin{itemize}
	\item \textbf{Diagnostic accuracy:} The average accuracy of the diagnostic object prediction.
    \item \textbf{Success rate:} The average number of epi\-sodes that end with a positive reward out of all episodes. In other words: the average ratio of episodes ended within the maximum amount of steps that did not end in a failure.
    \item \textbf{Episode length:} The average number of steps required for the completion of a level. At most, this is the maximum number of steps defined for each level.
    \item \textbf{Failure rate:} The ratio of episodes that end with the agent failing a task. Since there is a temporal ordering in the connectors, the agent is not allowed to visit them out of order. Similarly, if the agent fails to obey the twice/thrice modifier, the agent can fail the task by arriving at the next object too early.
    \item \textbf{Timeout rate:} The ratio of episodes that end without the agent completing the whole instruction, reaching the maximum number of steps.
\end{itemize}

Unless otherwise specified, we report the mean and standard devation over at least three different seeds to account for randomness factors in network initialization, the environment generation and the optimization process.

\subsection{Diagnostic training}
In this initial experiment, we add the diagnostic classifier to the agent (Aware model), and look at differences in how the training of the two models (Baseline and Aware) develop.
For the Aware model, we record also the diagnostic classifier's accuracy during training.

Furthermore, we perform an offline training test to check whether the hidden states of the agent are affected by diagnostic classification. Both the Baseline and Aware converged models are put in inference mode, and run for a fixed number of episodes. For all frames in these episodes, the hidden states and the correct diagnostic target are recorded. Together, they form an offline dataset, which we can use to train a new classifier, identical to the one used in the RL training. We then compare performance of the new classifier trained on the Baseline- vs. Aware-generated datasets.
%

\subsection{Source-level performance}
Here, we observe the performance of the two models on the levels they have been trained on.
Since the Aware model has an added task of making its hidden states explainable for a small classifier, convergence might take longer than the Baseline model. Furthermore, the base performance on the two novel complex levels can be examined using the source-level performance.

\subsection{Zero-shot generalization}
Next, we check whether the Aware agent can use the extra training signal for separating subtasks. By introducing an unseen characteristic to objects in the environment, the agent now has to identify which object it does know, and generalize learned behavior to the unknown object. Being able to isolate single objects in the environment should help the agent in this type of generealization.

Specifically, we consider the following cases:
\begin{itemize}
	\item \textbf{Color:} One object's color is replaced with an unknown color.
	\item \textbf{Object:} One object's type is replaced with an unknown type.
	\item \textbf{ColorObject:} One object's type and color are both replaced with unknowns.
\end{itemize}

In all cases, we only change a single object in the environment, such that the agent should be able to deduce which object is altered. This aids the agent in completing the given instruction, whereas changing multiple objects could lead to the agent visiting the objects in the wrong order more often.

\subsection{Sparse classification}
Lastly, an attempt is made to make the guiding signal more realistic. In the original setting, for every timestep in the environment we ask the agent for its current objective. However, in humans, intuitively this is too frequent, and should only be asked occasionally.

Therefore, we lower the frequency of the diagnostic classification. Instead of every frame, a classification is only asked up to a maximum of three times per game episode. Now, the extra signal is much lower, and both the classifier might need more time to reach convergence, as well as the feedback on the agent's hidden states is less dominant. We explore whether this is beneficial to the agent, considering the criteria from before. The training of the agent takes significantly longer, therefore only the \textsc{Before} and \textsc{Mixed-2} levels are considered.

\section{Results}
\label{sec:results}

Below, we give an overview of the results per experiment. First, the performance of the diagnostic classification task in general is presented. Second, all zero-shot experiments are shown. Finally, we elaborate on observations made during the sparse classification task.

\subsection{Diagnostic training}
\begin{figure}[]
    \centering
    \includegraphics[width=0.48\textwidth]{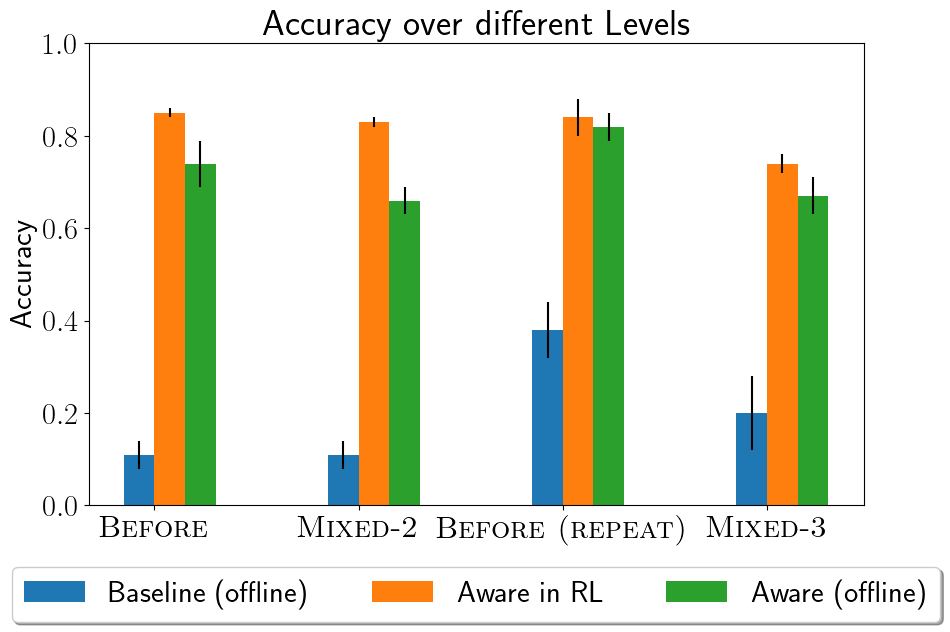}
    \caption{Diagnostic classification accuracy (and standard deviations) of the Baseline and Aware model on all levels.}
    \label{fig:results-interp}
\end{figure}
See Figure~\ref{fig:results-interp}. The Baseline model only has access to a classifier trained on the offline collected dataset, while the Aware model was evaluated at two different stages: once after training with RL, and once after retraining on the offline dataset.

Across all cases, the Aware model is able to predict the correct objective consistently. In the RL stage, the classifier is successfully trained, which indicates that the agent is still able to converge to a stable optimum. Furthermore, the subsequent difference between the offline  trained classifiers shows that the hidden states are positively affected by the training process: the Aware model's states are better suited for retraining the same classifier using a restricted dataset, and thus are more easily interpretable than the Baseline. Since this dataset is only a fraction of the number of frames that the agent observed during the RL stage, performance is slightly lower. Still, this shows that only a limited dataset is required before a classifier can be trained for the Aware model.

\subsection{Source-level performance}
\begin{table*}
    \caption{Performance of trained models in the source levels.}
    \label{tab:results-base}
    \centering
    \begin{tabular}{llclccc}
        \hline
                &               & \textbf{Frames} & \textbf{Episode length} & \textbf{Success rate} & \textbf{Fail rate}  & \textbf{Timeout}  \\ \hline
        Baseline    & \textsc{Before}         & 3k        & 10.9 ($\pm 0.1$)  & 1.00 ($\pm 0.00$) & 0.00 ($\pm 0.00$) & 0.00 ($\pm 0.00$) \\
                & \textsc{Mixed-2}   & 3k        & 10.6 ($\pm 0.1$)  & 1.00 ($\pm 0.01$) & 0.00 ($\pm 0.01$) & 0.00 ($\pm 0.00$) \\
                & \textsc{Before (repeat)}& 15k       & 50.1 ($\pm 8.4$) & 0.71 ($\pm 0.03$) & 0.08 ($\pm 0.03$) & 0.21 ($\pm 0.06$) \\
                & \textsc{Mixed-3}        & 6k        & 19.4 ($\pm 2.6$)  & 0.98 ($\pm 0.01$) & 0.02 ($\pm 0.01$) & 0.01 ($\pm 0.01$) \\
        Aware   & \textsc{Before}         & 3k        & 11.1 ($\pm 0.1$)  & 1.00 ($\pm 0.00$) & 0.00 ($\pm 0.00$) & 0.00 ($\pm 0.00$) \\
                & \textsc{Mixed-2}   & 3k        & 10.5 ($\pm 0.1$)  & 1.00 ($\pm 0.00$) & 0.00 ($\pm 0.00$) & 0.00 ($\pm 0.00$) \\
                & \textsc{Before (repeat)} & 11k       & 35.7 ($\pm 3.9$)  & 0.81 ($\pm 0.10$) & 0.10 ($\pm 0.05$) & 0.09 ($\pm 0.05$) \\
                & \textsc{Mixed-3}        & 5k        & 17.7 ($\pm 0.4$)  & 0.98 ($\pm 0.01$) & 0.02 ($\pm 0.01$) & 0.01 ($\pm 0.01$) \\ \hline
        \end{tabular}
\end{table*}
See Table~\ref{tab:results-base}. In the two most simple levels, there is little difference in the models, as both models agree on a seemingly optimal policy. However, on the complex levels, the two models show different behavior. Especially on the \textsc{Before (repeat)} level, the Aware model is able to reach a faster policy. Repeating a subtask is easier if the agent learns to disentangle objects better, and the increase in success rate shows that the Aware model is able to complete the compound instruction more often.

\begin{figure}[h]
    \centering
    \includegraphics[width=0.4\textwidth]{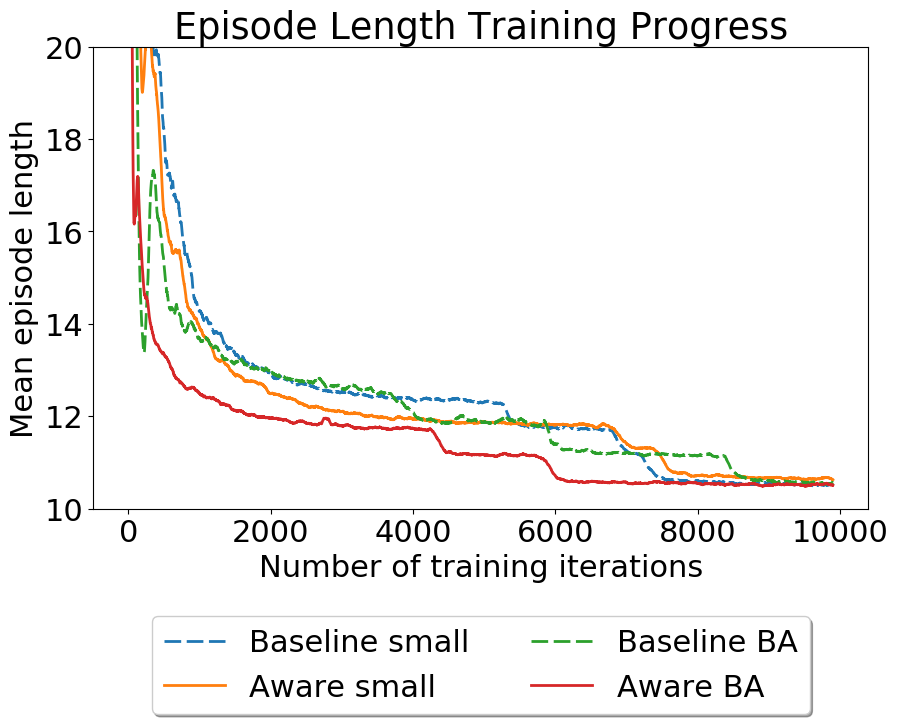}
    \caption{Training progress for the two models over the episode length metric, for two different levels. The dashed line indicates the Baseline, the solid indicates the Aware model. Each line is an average over multiple individual runs.}
    \label{fig:results-train-el}
\end{figure}

In Figure~\ref{fig:results-train-el}, the training progress can is plotted over the episode length metric for the two simple levels. Even though both models reach the same performance, there is a slight difference in their speed. Instead of the Aware model taking longer, because of the classification task, it can exploit the additional signal to learn slightly faster.

\subsection{Zero-shot generalization}
\begin{table*}
    \caption{Performance of a trained model on the source levels, applied in the new transfer learning setting. Here, there are three new scenarios: 1) a novel color, 2) a new type of object, 3) a combination of both.}
    \label{tab:results-zero}
    \centering
    \begin{tabular}{lllclc}
    \hline
     &  & \textbf{Transfer} & \textbf{Frames} & \textbf{Episode length} & \textbf{Success rate} \\
     \hline
     Base & \textsc{Before} & Color & 16k & 17.4 ($\pm 2.6$) & 0.76 ($\pm 0.06$) \\
     &  & Object & 53k & 57.8 ($\pm 14.1$) & 0.22 ($\pm 0.11$) \\
     &  & ColObj & 47k & 51.9 ($\pm 18.7$) & 0.31 ($\pm 0.12$) \\
     & \textsc{Mixed-2} & Color & 20k & 22.4 ($\pm 6.2$) & 0.67 ($\pm 0.06$) \\
     &  & Object & 70k & 77.6 ($\pm 12.7$) & 0.12 ($\pm 0.08$) \\
     &  & ColObj & 65k & 71.6 ($\pm 21.7$) & 0.14 ($\pm 0.12$) \\
     & \textsc{Before (repeat)} & Color & 54k & 59.9 ($\pm 14.4$) & 0.45 ($\pm 0.14$) \\
     &  & Object & 75k & 82.9 ($\pm 5.6$) & 0.17 ($\pm 0.10$) \\
     &  & ColObj & 69k & 75.6 ($\pm 4.3$) & 0.28 ($\pm 0.06$) \\
     & \textsc{Mixed-3} & Color & 40k & 44.0 ($\pm 10.0$) & 0.48 ($\pm 0.05$) \\
     &  & Object & 79k & 86.8 ($\pm 6.6$) & 0.07 ($\pm 0.04$) \\
     &  & ColObj & 64k & 70.4 ($\pm 10.0$) & 0.13 ($\pm 0.03$) \\
    Aware & \textsc{Before} & Color & 16k & 17.3 ($\pm 5.0$) & 0.77 ($\pm 0.09$) \\
     &  & Object & 49k & 54.4 ($\pm 12.8$) & 0.35 ($\pm 0.13$) \\
     &  & ColObj & 43k & 47.6 ($\pm 15.4$) & 0.41 ($\pm 0.10$) \\
     & \textsc{Mixed-2} & Color & 19k & 21.2 ($\pm 2.9$) & 0.68 ($\pm 0.03$) \\
     &  & Object & 53k & 58.6 ($\pm 4.1$) & 0.25 ($\pm 0.04$) \\
     &  & ColObj & 47k & 40.5 ($\pm 11.1$) & 0.40 ($\pm 0.07$) \\
     & \textsc{Before (repeat)} & Color & 45k & 49.4 ($\pm 6.7$) & 0.59 ($\pm 0.10$) \\
     &  & Object & 88k & 96.0 ($\pm 6.5$) & 0.10 ($\pm 0.06$) \\
     &  & ColObj & 89k & 97.4 ($\pm 7.9$) & 0.11 ($\pm 0.06$) \\
     & \textsc{Mixed-3} & Color & 41k & 45.1 ($\pm 9.6$) & 0.54 ($\pm 0.06$) \\
     &  & Object & 75k & 82.4 ($\pm 13.7$) & 0.24 ($\pm 0.05$) \\
     &  & ColObj & 61k & 67.5 ($\pm 8.2$) & 0.33 ($\pm 0.04$) \\
     \hline
    \end{tabular}
    \end{table*}

See Table~\ref{tab:results-zero}. In this case, we see an improvement for the Aware model in the two simple levels. Both in episode length and in success rate, the Aware model outperforms the baseline. Here, the \textsc{Mixed-2} level shows a larger difference than the easier \textsc{Before} level. This is is evidence for the need for complexity before the agent is able to exploit the language instruction fully.

However, for the complex levels, this difference is not as visible, but still the Aware model holds up to the baseline. When presented with only a new color, the Aware agent is able to be significantly faster, but in all other cases performance is comparable. Interestingly, the Aware model fails the whole instruction less often, but instead times out in both levels. This shift in termination reason is most likely due to the agent understanding that the known object in the level should not be visited yet, but fails to identify the unknown object. Upon inspection of the learned policy, the agent is actively avoiding the known object, but does not reach the other object in most cases. This shows that the training procedure did aid the agent in understanding its environment better: previously seen objects are more successfully identified, and the agent seems to know about their visiting order.

\subsection{Sparse classification}

\begin{figure}[]
    \centering
    \includegraphics[width=0.48\textwidth]{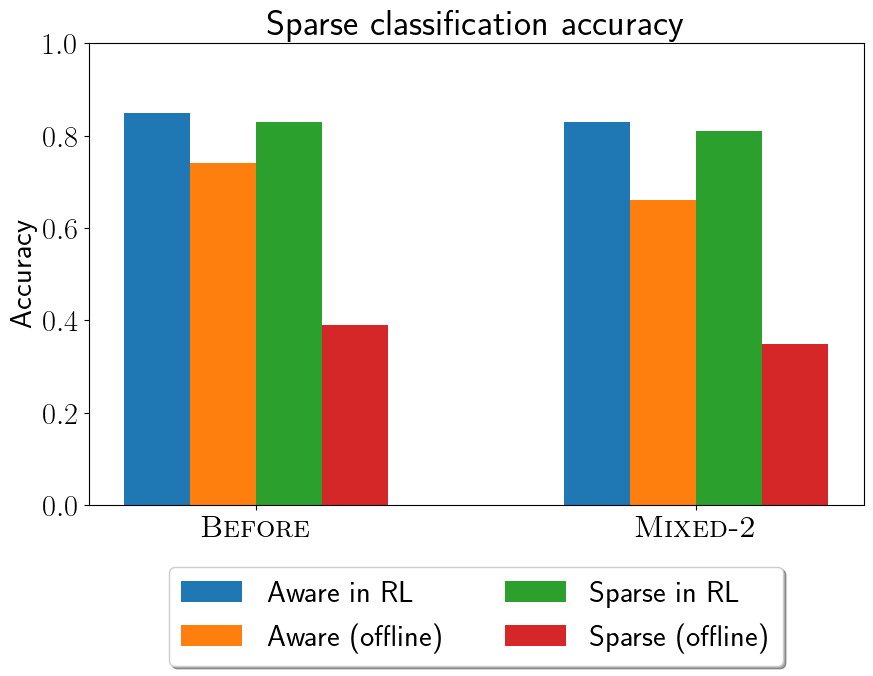}
    \caption{Diagnostic classficiation results for the standard and sparse versions of the Aware model. All values are averaged over at least two runs, with a standard deviation under 0.05.}
    \label{fig:results-sparse-interp}
\end{figure}

\begin{table}[tbp]
    \caption{Intra-level results for the standard and sparse versions of the Aware model. BA is the \textsc{Mixed-2} level. The policies learned by all agents were comparable and did not differ significantly over multiple runs.}
    \label{tab:results-sparse-base}
    \centering
    \begin{tabular}{llcc}
        \hline
        & & \textbf{EL} & \textbf{SR}\\
        \hline
        - & \textsc{Before} & 11.1 & 1.00 \\
        Sparse & \textsc{Before} & 10.9 & 1.00 \\
        - & \textsc{Mixed-2} & 10.6 & 1.00 \\
        Sparse & \textsc{Mixed-2} & 10.5 & 1.00 \\
        \hline
    \end{tabular}
\end{table}

\begin{table}
    \caption{Zero-shot performance of the sparsely trained models, compared to the standard Aware model.}
    \label{tab:results-sparse-zero}
    \centering
    \begin{tabular}{lllcc}
        \hline
        & & \textbf{Transfer} & \textbf{EL} & \textbf{SR}\\
        \hline
        - & \textsc{Before} & Color & 17.3 & 0.77 \\
        &  & Object & 54.4 & 0.35 \\
        &  & ColObj & 47.6 & 0.41 \\
        & \textsc{Mixed-2} & Color & 21.2 & 0.68 \\
        &  & Object & 58.6 & 0.25 \\
        &  & ColObj & 40.5 & 0.40 \\
        Sparse & \textsc{Before}&  Object & 16.1 &	0.81\\
        & & Object& 31.3 &	0.42\\
        & & ColObj& 24.7 &	0.52\\
        & \textsc{Mixed-2} & Color& 18.3 &	0.63\\
        & & Object  & 50.7 &	0.33\\
        & & ColObj& 41.0 &	0.43\\
        \hline
    \end{tabular}
\end{table}
Lastly, we present the results for the sparse diagnostic classification in Figure~\ref{fig:results-sparse-interp}, Table~\ref{tab:results-sparse-base} and Table~\ref{tab:results-sparse-zero}.

In comparison with the Baseline and Aware model, learning a policy for traversing the two simple levels does not take longer and reaches the same optimum as before (see Table~\ref{tab:results-sparse-base}). This is because the RL agent itself is unaffected by the changes in the classification procedure. However, the impact on the hidden states is considerably lower, as can be seen in Figure~\ref{fig:results-sparse-interp}. Here, the offline retrained classifier is not as easy to train as the standard Aware model. Still, compared to the earlier Baseline results, the Sparse classification is able to instigate some changes to the latent space.

In the zero shot experiments, there is some slight improvements in episode lengths and success rates. The hidden states may now be in balance between interpretability, as they can be organized by the retrained classifier to a certain degree, and efficiency, as the agent generalize them to unseen situations.

\section{Conclusion}

In this paper, we explored the addition of a simple classification task to a complex instruction-following RL problem. Through this addition, the agent was intended to become both more interpretable, and more aware of the compositional nature of the instructions. The results indicate that the agent is able to provide its current objective consistently, while having a minimal impact on the policy itself. Furthermore, these modified agents can be shown to be more general in zero-shot settings, suggesting that the added training signal helps in disentangling objects.

Future research should focus on expanding the level set that the agent was trained and evaluated on. Other types of instructions from the BabyAI environment, such as \textit{Pick up} or \textit{Put next} add more complexity to the task that the agents has to accomplish, and could also benefit from the improvements in object disentanglement. Additionally, adding obstacles such as separate rooms connected by doors, or distractor objects can interfere with the current setup. These situations form an interesting case for testing the diagnostic classification.

Finally, creating a more explicit hierarchical structure for the agent could make it more efficient in composing learned skills~\citep[e.g.][]{sutton1999between}. Such a hierarchical approach could use the training signal to train elementary skills and compose them more efficiently than in the current model.

\bibliography{main}
\bibliographystyle{acl_natbib_nourl}

\end{document}